\documentclass[runningheads]{llncs}
\usepackage[T1]{fontenc}
\usepackage{booktabs}
\usepackage{amsmath}
\usepackage{amsmath}
\usepackage{amssymb}
\usepackage{multirow}
\usepackage{url}
\usepackage{hyperref}
\usepackage{graphicx}
\usepackage{subcaption}
\usepackage[nocompress]{cite}

\usepackage{graphicx,verbatim}

\begin{document}
\title{Panda: Unsupervised Pelvic Anomaly Detection for Real-Time MR Imaging}

\author{Anika Knupfer\inst{1,2} \and Maximilian Lindholz\inst{3} \and Johanna Paula Müller\inst{4} \and Jordina Aviles Verdera\inst{1,2} \and Smiti Tripathy \inst{5} \and Susanne Schulz-Heise \inst{5} \and Jana Hutter\inst{1,2}}

\authorrunning{A. Knupfer et al.}

\institute{Institute of Information Processing, Leibniz University Hannover, Hannover, Germany 
\and CAIMED, L3S, Hannover, Germany\\
\and Department of Radiology, Charité Universitätsmedizin Berlin, Berlin, Germany\\
\and Image Data Exploration and Analysis Lab, Friedrich-Alexander University Erlangen-Nürnberg, Erlangen, Germany\\
\and Institute for Radiology, University Hospital Erlangen, Erlangen, Germany\\
\email{knupfer@tnt.uni-hannover.de}}

\maketitle              
\begin{abstract}
Female pelvic diseases remain an under researched area characterized by often delayed diagnosis. While pelvic MRI offers superior soft-tissue contrast for diagnosis and image-guided procedures, real-time anomaly detection remains challenging due to physiological motion, tissue deformation, and instrument artifacts. Existing supervised approaches are impractical, as adverse events are rare, heterogeneous, and difficult to annotate. We present a Dinomaly-based unsupervised anomaly detection framework adapted for pelvic MRI that learns normative representations from healthy cases and flags deviations without requiring labels. Our approach leverages a frozen DINOv3 Vision Transformer encoder combined with a noisy MLP bottleneck and Linear Attention decoder to prevent identity mapping while maintaining computational efficiency. Anomalies are localized via per-token cosine distance between encoder and decoder representations, yielding spatial anomaly maps that provide immediate feedback at the scanner to support radiologist decision-making and adaptive protocol adjustment. Evaluated on a curated subset of the Uterine Myoma Dataset, the framework achieves a pixel-level AUROC of 88.06\% and high specificity (95.45\%) at frame level at 40.5~slices/s, meeting real-time clinical deployment requirements. The spatial anomaly maps and frame-level scores provide immediate, localized feedback at the scanner to support radiologist decision-making and adaptive protocol adjustment during active procedures.


\end{abstract}

\section{Introduction}
Female pelvic diseases such as adenomyosis, endometriosis, cervical and endometrial cancer affect a significant number of women but remain underdiagnosed, with for example endometriosis affecting 10\% of women of reproductive age but suffering from an average time-till-diagnosis of 8 years. The complex anatomy, physiological motion, high variability and proximity of multiple organs are significant challenges both for diagnostic imaging and for increasingly available image-guided interventions~\cite{potter2021mri,moynagh2021image,yu2025application,white2019realizing}. MRI plays thereby a growing role~\cite{uka2024magnetic,kilbride2022mri} and is ideally suited particularly to guide interventions such as biopsies, ablations and brachytherapy in the anatomically complex and motion-prone pelvic region~\cite{romano2024uterus,wasnik2011normal}. However, despite these benefits carrying the promise of accelerating diagnosis and minimizing treatments side-effects, pelvic MRI remains challenging due to substantial physiological motion from respiration, peristalsis, and bladder filling~\cite{hernando2022quantitative} and the lack of interactivity in conventional MRI examinations~\cite{kilbride2022mri,liang2023mr}. The general paucity of well-curated data and labels in this under-researched area exhibiting substantial physiological variability further aggravates the problem~\cite{huang2023anomaly,bruno2015understanding}.\\

\noindent Classical supervised machine learning approaches are particularly ill-suited, as many clinically relevant anomalies are rare, heterogeneous, and difficult to label~\cite{rahmaniar2025multi}. Moreover, for interventional settings, it is impractical to collect large, well-annotated datasets representing adverse events~\cite{tajbakhsh2020embracing}. This motivates the use of unsupervised anomaly detection, where models learn a representation of “normal” pelvic anatomy and dynamics from routine cases and flag deviations without requiring explicit labels~\cite{knupfer2026unsupervisedanomalydetectiondiseases}. Recent advances in deep learning, spatio-temporal modeling, and generative models can potentially detect subtle deviations that may be invisible to the naked eye or that emerge gradually over time~\cite{pinaya2022fast,kustner2020cinenet}. However, translating these techniques to pelvic MRI poses unique challenges and requires due to the inherent data paucity bespoke methods~\cite{tajbakhsh2020embracing,knoll2020deep}. \\

\noindent This work contributes (1) a novel framework for unsupervised pelvic anomaly detection, systematically redesigning the Dinomaly architecture~\cite{Guo_2025_CVPR} for this domain, directly addressing the annotation bottleneck that renders supervised approaches impractical in this setting, (2) a reproducible evaluation benchmark for unsupervised pelvic MRI anomaly detection, and provide the first head-to-head comparison between reconstruction-based and generative detection paradigms, revealing complementary strengths that inform future system design and (3) clinically actionable real-time performance, delivering simultaneous frame-level slice flagging and spatially precise pixel-level anomaly maps immediately available at the scanner to adapt acquisition protocols immediately.

\section{Theory}
\label{sec:background}

\subsection{Problem Formulation}
\label{sec:problem}
We formulate anomaly detection as an unsupervised learning problem. $\mathcal{D}_{\text{train}} = \{x_i\}_{i=1}^{N}$ are the set of $N$ training frames extracted from procedurally normal pelvic MRI sequences, acquired in the absence of pathologies, major artifacts, adverse events and clinically significant deviations. The model learns a representation of normal pelvic appearance, such that deviating inputs induce elevated reconstruction error. Anomaly localization is performed via per-pixel cosine dissimilarity between encoder and decoder representations. Each frame $x \in \mathbb{R}^{H \times W}$ is a 2D sagittal slice, and the model produces a pixel-level anomaly map $\mathcal{A}(x) \in [0,1]^{H \times W}$, where $\mathcal{A}(x)_{h,w}$ quantifies local deviation from normality at position $(h,w)$ for $h \in \{1,\ldots,H\}$, $w \in \{1,\ldots,W\}$. The frame-level anomaly score $s(x) = \max_{h,w}\,\mathcal{A}(x)_{h,w} \in [0,1]$ captures the most pronounced anomalous responses within the frame, providing a global slice classification. Inference operates under near real-time constraints.

\subsection{Dinomaly-Based Architecture}
\label{sec:overview}
Our approach extends Dinomaly~\cite{Guo_2025_CVPR}, a minimalist reconstruction-based anomaly detection architecture built exclusively from Transformer blocks~\cite{vaswani2017attention}. Operating at the representation level improves robustness to the pronounced intensity variability typical for MR imaging. The encoder is a pre-trained foundation model, a Vision Transformer (ViT). It remains entirely frozen during training and inference, with feature maps from intermediate layers extracted as reconstruction targets. Freezing prevents encoder-decoder co-adaptation and preserves the pretrained representation space as a stable normative reference. The bottleneck is a MLP with training-time Dropout that introduces stochastic feature corruption, limiting the decoder's generalization to unseen inputs to avoid the reconstruction of anomalous patterns it was never trained on~\cite{you2022unified}. By corrupting aggregated encoder features during training, the bottleneck forces the decoder to restore normal structure from incomplete information, ensuring elevated reconstruction error on anomalous inputs at inference. Dropout is disabled at inference for deterministic operation. The decoder replaces standard Softmax attention with Linear Attention (LA)~\cite{katharopoulos2020transformers} to promote globally distributed feature aggregation over spatially concentrated token interactions, well-suited to pelvic MRI where anomalies manifest as subtle contextual inconsistencies. LA is defined as
\begin{equation}
  \operatorname{LA}(Q, K, V) = \phi(Q)\bigl(\phi(K)^{\top} V\bigr),
  \quad \phi(\cdot) = \operatorname{elu}(\cdot) + 1,
\end{equation}
where $Q, K, V$ are the query, key, and value matrices, and $\phi(\cdot)$ is an element-wise non-negative activation enabling the kernel factorization, favoring globally consistent reconstruction over strict token-wise correspondence. Training minimizes a group-to-group cosine reconstruction loss. Encoder feature maps are partitioned into two groups capturing complementary levels of abstraction: low-level features (layers~3-6), encoding local texture and high-level features (layers~7-10), encoding semantic context. The decoder is trained to reconstruct each group independently, with gradient shrinking to emphasize hard tokens and sharpen sensitivity to subtle deviations. Denoting the flattened grouped encoder and decoder feature maps as $F(f_{E})$ and $F(\hat{f}_{D})$, the loss is
\begin{equation}
  \mathcal{L} = D_{\cos}\!\bigl(F(f_{E}),\, F(\hat{f}_{D})\bigr),
  \quad D_{\cos}(a, b) = 1 - \frac{a^{\top} b}{\|a\|\,\|b\|},
  \label{eq:loss}
\end{equation}

\section{Method}
\label{sec:method}

\begin{figure}[!t]
\centering
\begin{subfigure}[t]{0.46\columnwidth}
    \centering
    \includegraphics[width=\linewidth]{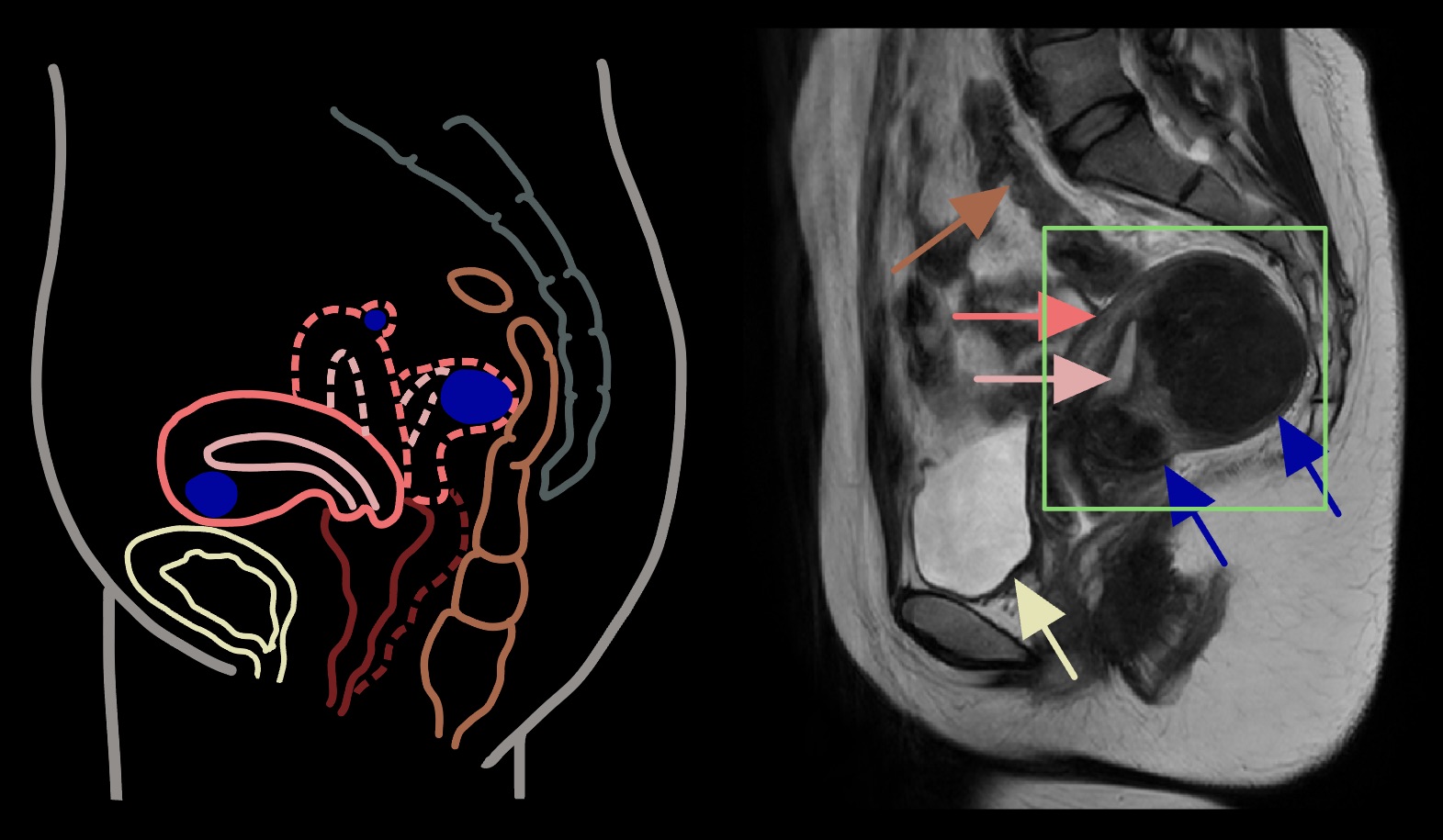}
    \caption{Anatomy.}
    \label{fig:anatomy}
\end{subfigure}
\hfill
\begin{subfigure}[t]{0.50\columnwidth}
    \centering
    \includegraphics[width=\linewidth]{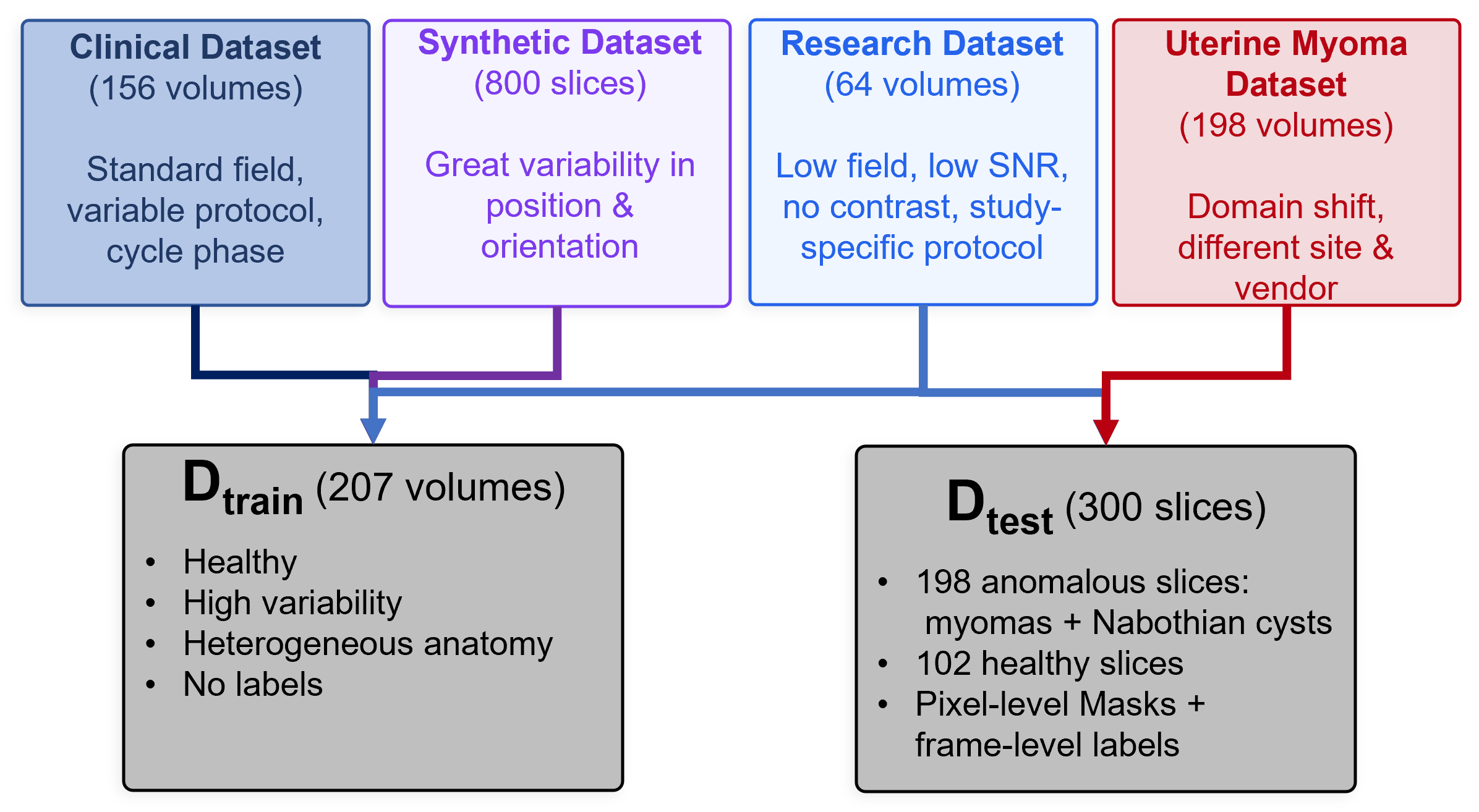}
    \caption{Datasets.}
    \label{fig:datasets}
\end{subfigure}
\caption{(a)~Schematic and T2w MRI pelvic illustration (pink—uterus; light pink—endometrium; blue—myoma; yellow—bladder; brown—rectum. Dashed lines indicate physiological uterine variability and the green bounding box the uterine crop. (b)~Data flowchart illustrating the assembly of $\mathcal{D}_{\text{train}}$ and $\mathcal{D}_{\text{test}}$.}
\label{fig:anatomy_data}
\end{figure}

\subsection{Datasets}
\label{sec:data}
\paragraph{Training Data.}
$\mathcal{D}_{\text{train}}$ comprises 207 anonymized sagittal T2w healthy MRI volumes (see Fig.~\ref{fig:anatomy}, ethics numbers: 24-304-Br, 23-444-Bm), assembled from a clinical and a research dataset supplemented with synthetic data (Fig.~\ref{fig:datasets}). The clinical dataset with 156 volumes, employed with institutional ethics approval exhibits substantial field strength, protocol, usage of contrast and physiological variability. The research dataset, consisting of 51 volumes, was acquired at low field (0.55T) and is hence characterized by low SNR. The 800 publicly available synthetic images~\cite{zenodoUterus} covering especially rare anatomical variants were reviewed by radiologists for anatomical plausibility and verified against existing training samples using cosine similarity (threshold 0.95) to avoid near-duplicate images.

\paragraph{Evaluation Data and out-of-distribution evaluation.}
$\mathcal{D}_{\text{test}}$ combines 198 anomalous volumes from the public Uterine Myoma Dataset (UMD)~\cite{pan2024large,pan2023} and 13 low field research dataset volumes held out for frame-level evaluation. The UMD dataset was acquired at a different site with a different vendor and greatly varying parameters and thus constitutes a genuine out-of-distribution evaluation.

\paragraph{Data Preparation.}
All volumes were resampled to $0.5{\times}0.5{\times}1$~mm$^3$, resized to $256{\times}256{\times}30$, and intensity-normalized to $[0,1]$. For $\mathcal{D}_{\text{train}}$, masking was performed with an attention-based 3D U-Net~\cite{tripathydeep}, for $\mathcal{D}_{\text{test}}$ provided masks were used. Bounding boxes of size $96{\times}96{\times}n$ ($n$~=~slices containing uterine tissue) were extracted (See Fig.~\ref{fig:anatomy}). Synthetic training images required no further spatial preprocessing. For evaluation, one slice per examination containing both annotated anomalous regions and visible uterine structures was selected, yielding 198 anomalous slices. The 102 anomaly-free slices were sampled at a minimum distance of 2 slices from any annotated lesion (UMD) or with a minimum inter-slice spacing of 2 (held-out volumes) to avoid similar neighboring frames.

\subsection{Architecture and Training.}
\begin{figure}[!t]
\centering
\centerline{\includegraphics[width=\columnwidth]{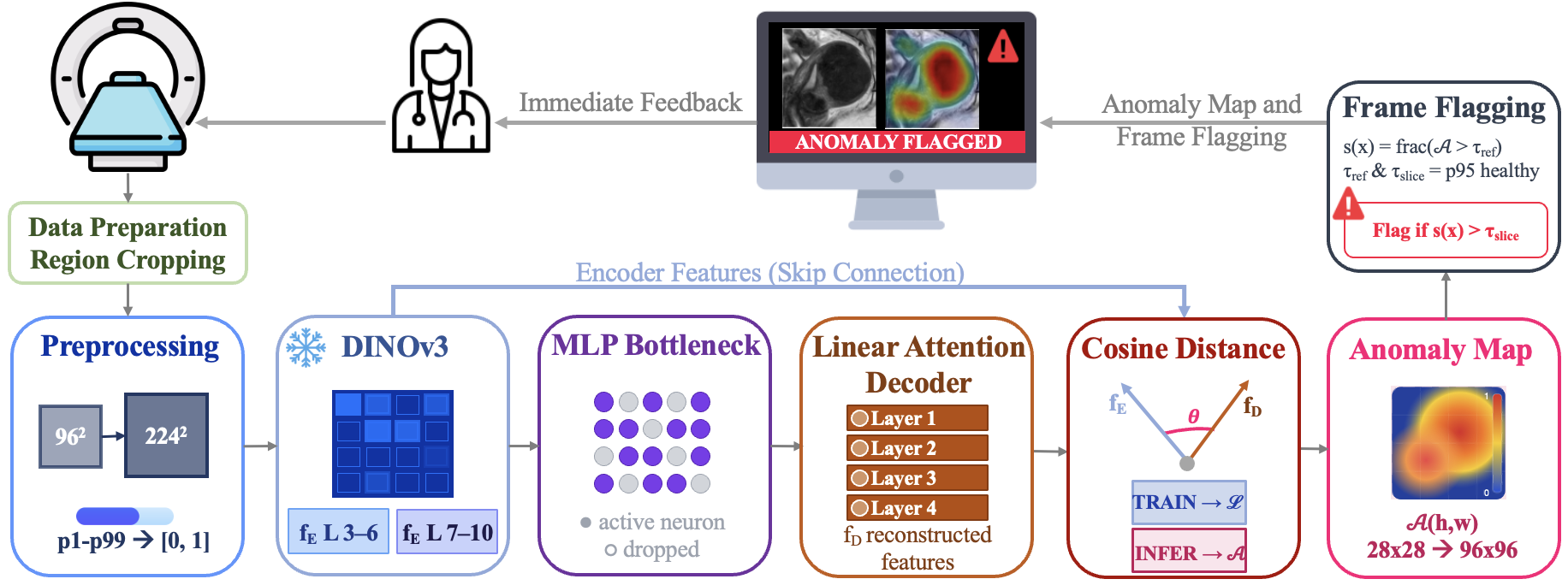}}
\caption{Proposed framework for low-latency pelvic anomaly detection in MRI.}
\label{fig:architecture}
\end{figure}

The encoder is a frozen DINOv3 ViT-L/16~\cite{simeoni2025dinov3,hf_dinov3}, selected to balance representational capacity and overfitting risk in a low-data setting. Input slices are resized from $96{\times}96$ to $224{\times}224$ and normalized using per-slice percentile scaling to $[0,1]$ (1st-99th percentile) followed by standard ImageNet normalization, producing patch-level embeddings of dimension $d=1024$ on a $28{\times}28$ grid. The bottleneck MLP applies dropout ($p=0.136$) during training, and the decoder consists of four Linear Attention layers with eight attention heads and an MLP ratio of $2.097$. The anomaly map $\mathcal{A}(h,w) = D_{\cos}(f_E, \hat{f}_D)$ is computed via a skip connection bypassing the bottleneck and upsampled from $28{\times}28$ to $96{\times}96$ for evaluation. The model is trained for 120 epochs with batch size 32, learning rate $4.24{\times}10^{-4}$, and weight decay $10^{-4}$ using AdamW, without data augmentation. Hard-token weighting via gradient shrinking is applied from the first epoch at the 33rd percentile of per-token reconstruction error, down-weighting well-reconstructed tokens by a factor of $0.1$. Hyperparameters were selected via Bayesian optimization over 40 configurations directly maximizing pixel-level AUROC on $\mathcal{D}_{\text{test}}$, with no gradient updates derived from test data, as validation loss showed negligible correlation with detection performance ($\rho_S = +0.094$, $p = 0.566$). The full pipeline is illustrated in Fig.~\ref{fig:architecture} including the in-built feedback mechanism available for instant feedback at the time of acquisition.

\subsection{Evaluation.}
\textbf{Pixel-Level Anomaly Scoring} is performed on the continuous anomaly maps $\mathcal{A}(h,w)$ using AUROC. For binary metrics, a pixel threshold $\tau_{\text{px}}$ is determined by maximizing the Dice coefficient across $\mathcal{D}_{\text{test}}$, providing an upper bound on performance. 
\textbf{Frame-Level Anomaly Scoring} reduces the anomaly map $\mathcal{A}$ to a scalar score measuring the spatial extent of anomalous responses to capture the prevalence of atypical responses, avoiding the noise sensitivity of maximum-based scores and the dilution of localized anomalies in mean-based scores:
\begin{equation}
  s(x) = \frac{1}{HW} \sum_{h,w}
  \mathbf{1}\bigl[\mathcal{A}(h,w) > \tau_{\text{ref}}\bigr].
\end{equation}
Both $\tau_{\text{ref}}$ and the slice-level decision threshold $\tau_{\text{slice}}$ are calibrated exclusively on healthy slices from $\mathcal{D}_{\text{train}}$. $\tau_{\text{ref}}$ is set to the 95th percentile of pixel-level anomaly scores and $\tau_{\text{slice}}$ to the 95th percentile of $s(x)$. A frame is flagged as anomalous at inference if $s(x) > \tau_{\text{slice}}$.

\section{Experiments and Results}
\label{sec:experiments}
\noindent We evaluate the proposed method on $\mathcal{D}_{\text{test}}$ at pixel-level and frame-level, and benchmark against a current state of the art ResVAE~\cite{knupfer2026unsupervisedanomalydetectiondiseases}.

\begin{figure}[!t]
\centering
\centerline{\includegraphics[width=0.7\columnwidth]{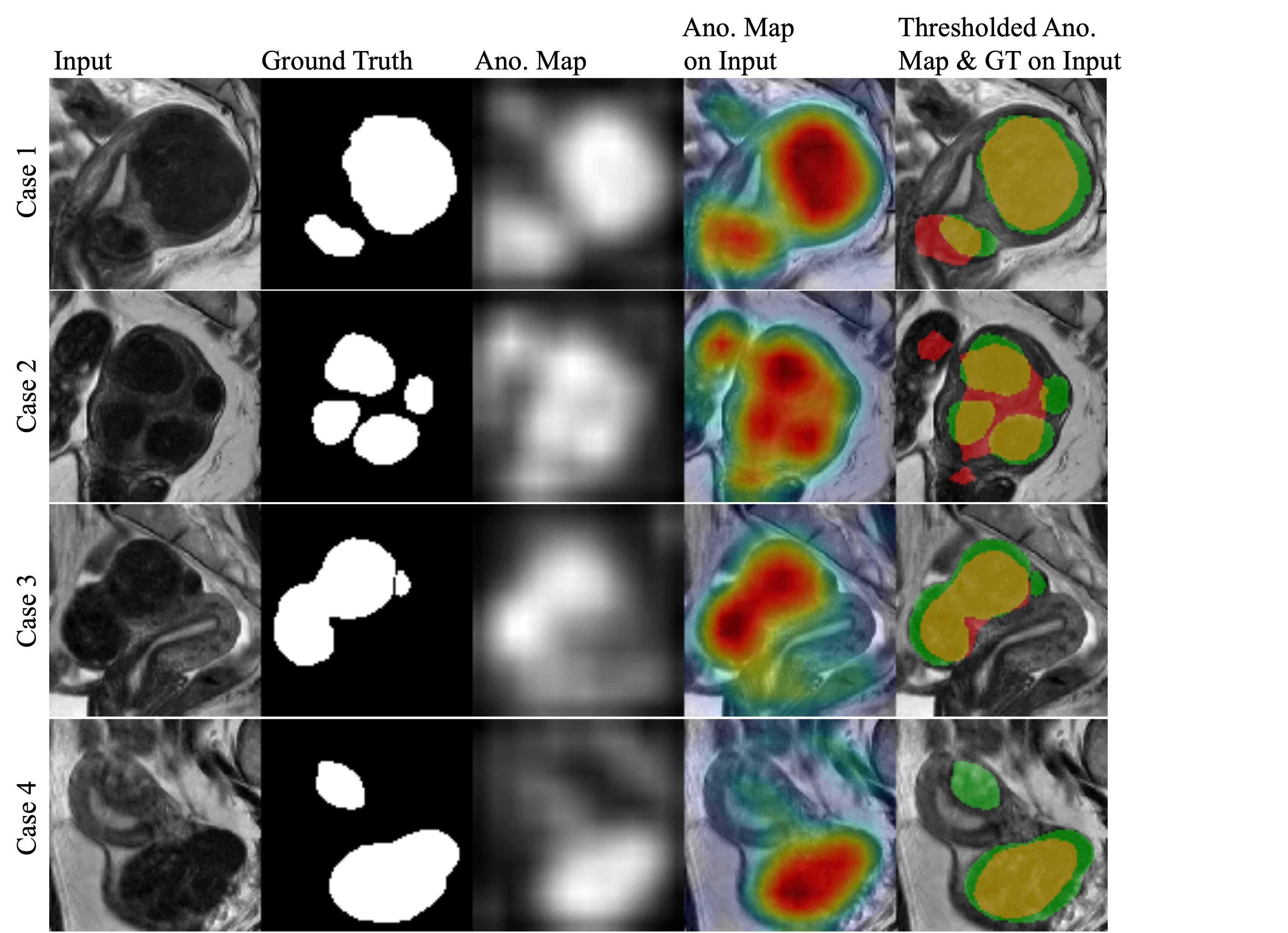}}
\caption{Representative results on $\mathcal{D}_{\text{test}}$. Each row presents from left to right: T2w input slice, GT, raw anomaly map, heatmap overlay on the input, and threshold-based prediction maximizing the Dice across $\mathcal{D}_{\text{test}}$ overlaid with GT on the input (yellow: predicted anomaly area, green: GT boundary, red: false positive).}
\label{fig:qualitative}
  \end{figure}

\paragraph{Anomaly Detection Performance.}
Qualitatively, 4 representative examples (Fig.~\ref{fig:qualitative}) show anomaly maps corresponding closely with GT. Two failure patterns emerge with smaller or low-contrast lesions (case 4) and adjacent structures like the bowel occasionally causing false positives (case 2).
Results comparing upscaled $96{\times}96$ anomaly maps with expert-annotated segmentation masks are listed in Table~\ref{tab:results}. The pixel-level AUROC of 88.06\% demonstrates strong detection ability. At the retrospectively determined optimal threshold, accuracy reaches 93.62\% and specificity 95.10\%, demonstrating that the vast majority of normal pixels are classified correctly. The low sensitivity of 38.49\% and precision of 17.31\% is linked to the strong class imbalance: pathological areas occupying only a small fraction of pixels results in even spatially well-localized anomaly maps demonstrating relatively low precision and sensitivity on a pixel-by-pixel basis. The F1 score of 23.87\% should be interpreted in this context. Taken together, the high AUROC value and high specificity suggest that the model generates spatially meaningful anomaly maps, while the precise delineation of lesion boundaries remains challenging. In frame-level results, the healthy-calibrated threshold $\tau_{\text{slice}}$ yields high specificity (95.45\%) and precision (71.88\%) (Table~\ref{tab:results}). The low sensitivity (11.61\%) is a direct consequence of the conservative threshold and the coarse token resolution where small lesions may not elevate a sufficient fraction of pixels to trigger a flag. The frame-level AUROC of 68.41\% confirms moderate discriminative ability, and the high-specificity operating point reflects a design choice prioritizing trust over exhaustive recall. Per-slice inference latency, measured on the UMD test set using an NVIDIA GeForce RTX 3080 (10\,GB), was 0.0247\,s per slice or 40.5\,slices/s, consistent with real-time requirements for MRI workflows.

\begin{table}[t]
\centering
\scriptsize
\caption{Detection performance (\%) on $\mathcal{D}_{\text{test}}$, comparing the proposed Dinomaly-based architecture with the state-of-the-art ResVAE~\cite{knupfer2026unsupervisedanomalydetectiondiseases}. Best scores in bold.}
\label{tab:results}
\resizebox{\columnwidth}{!}{
\begin{tabular}{llcccccc}
\toprule
\textbf{Architecture}&\textbf{Evaluation } & \textbf{AUROC~$\uparrow$} & \textbf{Accuracy~$\uparrow$} & \textbf{Precision~$\uparrow$} & \textbf{Sensitivity~$\uparrow$} & \textbf{Specificity~$\uparrow$} &\textbf{F1~$\uparrow$}  \\
\midrule
\multirow{2}{*}{Dinomaly-based} & Frame-level & 68.41 & 53.53 & 71.88 & 11.61 & 95.45 & - \\
& Pixel-level & \textbf{88.06} & \textbf{93.62} & \textbf{17.31} & 38.49 & \textbf{95.10} & \textbf{23.87} \\
\midrule
SotA ResVAE~\cite{knupfer2026unsupervisedanomalydetectiondiseases}& Pixel-level & 69.65 & 64.63 & 6.95 & \textbf{79.91} & 64.43 & 11.21\\
\bottomrule
\end{tabular}
}
\end{table}

\begin{figure}[t]
\centering
\centerline{\includegraphics[width=\columnwidth]{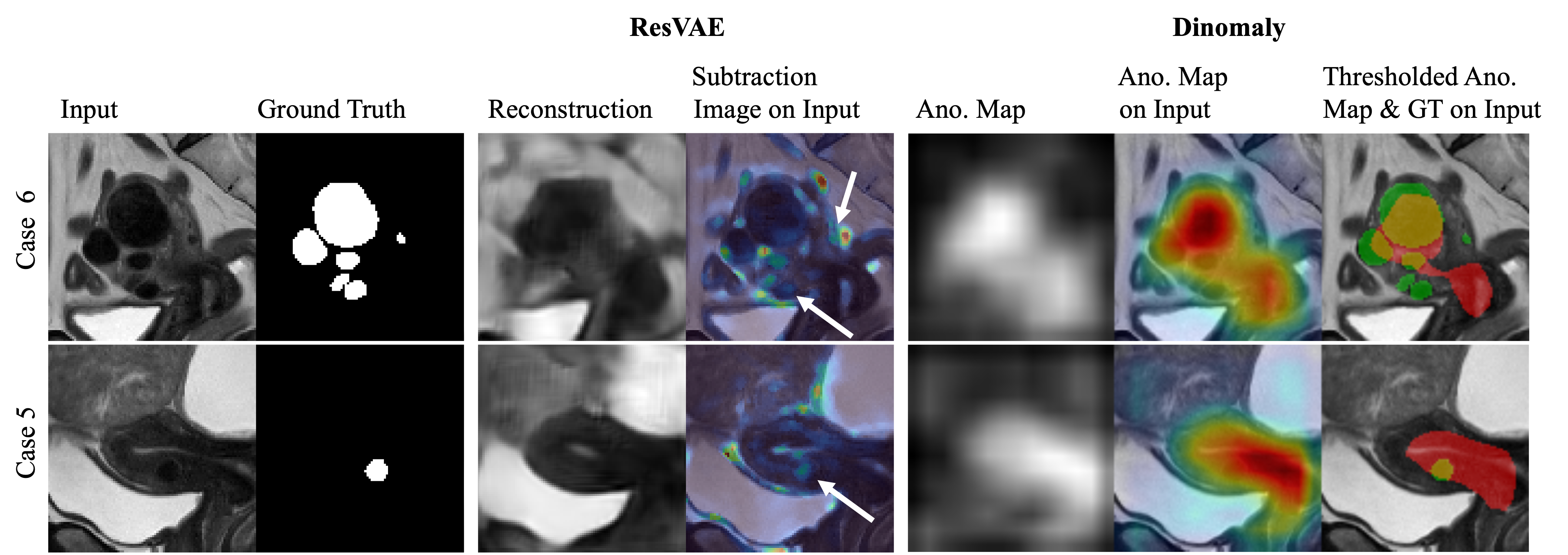}}
\caption{Qualitative comparison on 2 representative cases: Input slice, GT mask, ResVAE reconstruction and subtraction heatmap overlay, Dinomaly anomaly map, heatmap overlay, and thresholded prediction with GT boundary (yellow: true positive, green: GT, red: false positive). Arrows indicate subtle lesions.}
\label{fig:VAE_Dino}
\end{figure}

\paragraph{Comparison with ResVAE Baseline.}
The re-trained ResVAE~\cite{knupfer2026unsupervisedanomalydetectiondiseases} achieves markedly lower pixel-level AUROC (69.65 vs.\ 87.85) and precision (6.95 vs.\ 17.04), but substantially higher sensitivity (79.91 vs.\ 38.39), reflecting its tendency to produce diffuse, high-recall anomaly maps at the cost of poor specificity (64.43 vs.\ 95.02). Figure~\ref{fig:VAE_Dino} illustrates the complementary nature of the two approaches across two representative cases: the reconstruction-based ResVAE method responds to subtle low-contrast lesions with greater sensitivity (white arrows) while Dinomaly provides spatially precise localization of dominant anomalous structures, consistent with the semantic representations of the frozen DINOv3 encoder. The coarser $28{\times}28$ token grid remains the primary limitation of the proposed method with respect to small lesion sensitivity.

\begin{table}[t]
\centering
\caption{Ablation study results (\%) on pixel-level performance. All modifications are evaluated relative to the full model (Sec.~\ref{sec:method}). Metric scores above Base model scores are highlighted in bold.}
\label{tab:ablation}
\resizebox{\columnwidth}{!}{
\begin{tabular}{llccccc}
\toprule
\textbf{Group} & \textbf{Configuration} & \textbf{AUROC~$\uparrow$} & \textbf{Accuracy~$\uparrow$} & \textbf{Precision~$\uparrow$} & \textbf{Sensitivity~$\uparrow$} & \textbf{Specificity~$\uparrow$} \\
\midrule
Base & Full model & 88.06 & 93.62 & 17.31 & 38.49 & 95.10 \\
\midrule
\multirow{4}{*}{Architectural} 
& ViT-S Backbone & 83.77 & 90.67 & 16.63 & \textbf{42.7} & 92.38 \\
& 12 Decoder Layers & 87.75 & 93.00 & 16.34 & \textbf{41.17} & 94.38 \\
& 8 Attention Heads & 87.02 & 92.27 & 19.72 & \textbf{40.58} & 94.11 \\
& 0 Decoder MLP Ratio & 88.04 & \textbf{93.88} & \textbf{17.60} & 36.86& \textbf{95.40} \\
\midrule
\multirow{3}{*}{Training} 
& 2 Augmentations/slice & 87.23 & 92.97 & 15.98 & \textbf{40.04} & 94.11 \\
& No Synthetic Data & 87.81 & 93.36 & 16.77 & \textbf{39.27} & 94.80 \\
& No Dropout Bottleneck & \textbf{88.12} & \textbf{93.70} & \textbf{17.47} & 38.27 & \textbf{95.18} \\
\midrule
Data scaling 
& 50\% of $\mathcal{D}_{\text{train}}$ & 83.76 & 92.01 & 13.16 & 37.08 & 93.47 \\
\bottomrule
\end{tabular}
}
\end{table}

\paragraph{Ablation Study.}
The small differences between models (Table~\ref{tab:ablation}) reflect the robustness of the frozen DINOv3 encoder, whose representations are largely insensitive to reconstruction decisions. Individual modifications involve trade-offs (e.g., removing dropouts improves accuracy/specificity, while ViT-S/16 and larger decoders increase sensitivity but carry the risk of overfitting). The full configuration was selected based on the best overall balance. Replacing ViT-L/16 with ViT-S/16 reduces the AUROC by 4.1\% but significantly lowers inference latency (76.8\,slices/s). Training with 50\% results in significant deterioration.

\section{Discussion and Conclusion}
We presented a systematic adaptation of the Dinomaly framework for unsupervised anomaly detection in pelvic T2w MRI, leveraging a frozen DINOv3 ViT-L/16 encoder with LA decoding and a noisy MLP bottleneck. The strong pixel-level AUROC (88.06\%) is largely attributable to the frozen encoder, whose semantic representations generalize well to pelvic MRI and prevent co-adaptation with the decoder, preserving a stable normative reference. The ablation study confirms robustness in a data-scarce setting. At frame level, sensitivity (11.61\%) is deliberately sacrificed for high specificity (95.45\%), the conservative healthy-calibrated threshold minimizes false alarms, as spurious flagging would erode clinical trust. The coarse $28{\times}28$ token grid is the primary limitation, causing small or low-contrast lesions to produce insufficient anomalous responses. The complementary VAE baseline, high sensitivity (79.91\%) but low specificity (64.43\%), highlights the benefits of potentially fusing both paradigms. Clinically, the system is designed as a real-time decision support tool: when a frame is flagged, the spatial anomaly map is immediately available at the scanner, enabling the radiologist to adapt. Generalization beyond myomas and Nabothian cysts, to other pathologies and interventional settings, remains to be validated. 

\begin{credits}
\subsubsection{Acknowledgement} 
This work was supported by the Bavarian State Ministry of Health, Care and Prevention [project EndoKI], DFG Heisenberg [502024488], ERC StG EARTHWORM [101165242], ERC Proof-of-concept grant SYNCWORM [101293293] and CAIMed - Lower Saxony Center for Artificial Intelligence and Causal Methods in Medicine [ZN4257] funding. The authors have no competing interests to declare.
\end{credits}

\bibliographystyle{splncs04}
\bibliography{refs}

\end{document}